\documentclass[preprint,12pt]{elsarticle}

\usepackage{amssymb}
\usepackage{amsmath}
\usepackage{hyperref}
\usepackage{booktabs}
\usepackage{listings}
\usepackage{algorithm}
\usepackage{algpseudocode}

\journal{Neurocomputing}

\begin{document}

\begin{frontmatter}



\title{MoIRA: Modular Instruction Routing Architecture for Multi-Task Robotics} 


\author[label1]{Dmytro Kuzmenko}
\author[label2]{Nadiya Shvai}

\affiliation[label1]{organization={Department of Multimedia Systems, National University of Kyiv-Mohyla Academy},
            addressline={Hryhoriya Skovorody St, 2}, 
            city={Kyiv},
            postcode={04655}, 
            country={Ukraine}}


\affiliation[label2]{organization={Department of Mathematics, National University of Kyiv-Mohyla Academy},
            addressline={Hryhoriya Skovorody St, 2}, 
            city={Kyiv},
            postcode={04655}, 
            country={Ukraine}}

\begin{abstract}

Mixture-of-Experts (MoE) approaches have gained traction in robotics for their ability to dynamically allocate resources and specialize sub-networks. However, such systems typically rely on monolithic architectures with rigid, learned internal routing, which prevents selective expert customization and necessitates expensive joint training. We propose MoIRA, an architecture-agnostic modular framework that coordinates decoupled experts via an external, zero-shot text router. MoIRA employs two routing strategies: embedding-based similarity and prompt-driven language model inference. Leveraging Gr00t-N1 and $\pi_{0}$ Vision-Language-Action models with low-rank adapters, we evaluate MoIRA on GR1 Humanoid tasks and LIBERO benchmarks. Our approach consistently outperforms generalist models and competes with fully trained MoE pipelines. Furthermore, we demonstrate system robustness against instruction perturbations. By relying on textual descriptions for zero-shot orchestration, MoIRA proves the viability of modular deployment and offers a scalable, flexible foundation for multi-expert robotic systems.

\end{abstract}


\end{frontmatter}



\begingroup
\renewcommand\thefootnote{}
\footnotetext{This manuscript is the author-accepted version of an article published in \emph{Neurocomputing}, vol.~674, 132962 (2026). The final version of record is available at \url{https://doi.org/10.1016/j.neucom.2026.132962}.}
\endgroup


\begingroup
\let\clearpage\relax
\section{Introduction}

Robotic manipulation and navigation tasks have traditionally been tackled using reinforcement learning (RL) and imitation learning (IL) pipelines. These approaches have demonstrated strong performance across various settings but often depend on dense reward signals, curated expert demonstrations, or extensive task-specific tuning \cite{sharma2020emergentrealworldroboticskills, chen2021efficientlytrainingonpolicyactorcritic, liang2024neverendingbehaviorcloningagentrobotic}. Transformer-based models such as ACT \cite{zhao2023learningfinegrainedbimanualmanipulation} have further advanced robotic policy learning by enabling fine-grained, sequence-aware control.

More recently, foundation models have emerged as an alternative to traditional RL/IL pipelines, offering general-purpose capabilities without the need for task-specific training or reward engineering. Vision-language models (VLMs) such as PaLI-Gemma \cite{steiner2024paligemma2familyversatile}, LLaVA \cite{liu2023visualinstructiontuning}, and Qwen-VL \cite{bai2023qwenvlversatilevisionlanguagemodel} exhibit strong image-text grounding and instruction understanding. Although not designed for robotics, they can interpret natural language commands and scene context, making them useful for high-level planning and zero-shot inference.

Building on this trend, vision-language-action (VLA) models combine vision-language encoders with visiomotor control heads to support end-to-end robotic control. Recent examples include RT-2 \cite{brohan2023rt2visionlanguageactionmodelstransfer}, the RT-X family \cite{embodimentcollaboration2024openxembodimentroboticlearning}, OpenVLA \cite{kim24openvla}, MiniVLA \cite{belkhale2024minivla}, $\pi_0$ by Physical Intelligence \cite{black2024pi0visionlanguageactionflowmodel}, and Gr00t-N1 by NVIDIA \cite{nvidia2025gr00tn1openfoundation}. These models are typically pretrained on large-scale, diverse datasets (e.g. Open-X Embodiment \cite{embodimentcollaboration2024openxembodimentroboticlearning}) and incorporate heterogeneous data sources, including web-scale multimodal content, subtask annotations, and demonstrations from different robot embodiments. Their goal is to generalize across embodiments, task semantics, and modalities with minimal finetuning. However, their generalist nature can lead to reduced precision, inefficient memory usage, and difficulty scaling to large task libraries \cite{guruprasad2024benchmarking}.

\begin{figure}[htpb] 
  \centering
  \includegraphics[width=1.0\linewidth]{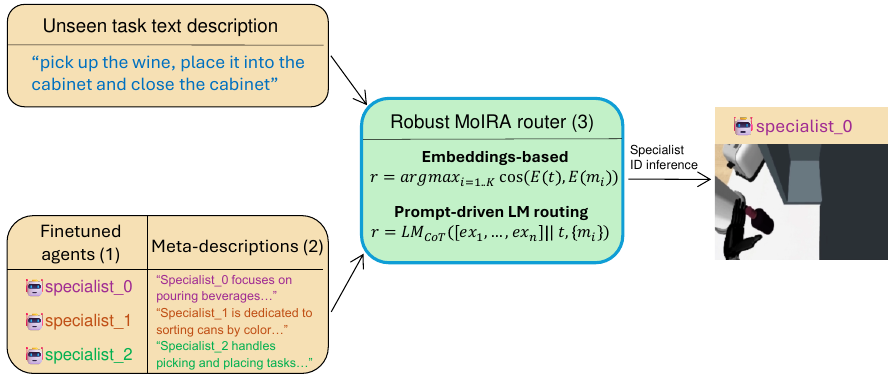}
      \caption{Overview of the MoIRA framework. The architecture decouples policy learning from task assignment by coordinating (1) a pool of independently \textbf{fine-tuned specialist agents} via (2) \textbf{textual meta-descriptions}. The (3) \textbf{Router core} performs zero-shot assignment of the unseen task $t$ to the optimal specialist index $r$ using either embedding similarity or prompt-driven reasoning. This design allows modular expert addition without retraining the routing mechanism.}
  \label{fig:moira_router_figure_1}
\end{figure}

Concurrently, Mixture-of-Experts (MoE) architectures have regained interest as a modular approach to specialization. Originally proposed by Jacobs et al. \cite{6797059} for adaptive task decomposition, MoEs are now widely explored for improving efficiency in large language models \cite{Cai_2025} and in robotic learning. In dexterous manipulation, residual MoE structures \cite{huang2024efficientresiduallearningmixtureofexperts} enable the composition of expert policies without relying on language inputs or transformers. MoLe-VLA \cite{zhang2025molevladynamiclayerskippingvision} introduces a spatial-temporal aware router to dynamically activate sublayers in VLA models based on token relevance. In the locomotion domain, MoRE \cite{zhao2025moreunlockingscalabilityreinforcement} and GERM \cite{song2024germgeneralistroboticmodel} apply sparse expert activation to quadruped control, balancing sample efficiency with generalization. Similarly, recent inference-optimization works like MoDE \cite{reuss2024efficientdiffusiontransformerpolicies} and Tra-MoE \cite{yang2025tramoelearningtrajectoryprediction} leverage sparsely-gated transformers to improve multi-domain policy prediction.

While these systems show promise, they share key limitations related to their architectural rigidity. Generalist models trained on broad data tend to suffer from degraded per-task performance and high runtime memory usage. Conversely, existing expert-based MoEs typically require orchestrated training and internal routing mechanisms tied to specific monolithic model structures. This introduces a critical tradeoff between specialization, modularity, and deployment flexibility that has not been fully addressed.

To address these limitations, we adopt a modular architectural perspective wherein each expert can be independently developed, customized, and optimized. This decouples training from deployment and allows for flexible reuse across tasks and embodiments. We therefore propose \textbf{Mo}dular \textbf{I}nstruction \textbf{R}outing \textbf{A}rchitecture, \textbf{MoIRA} (Figure \ref{fig:moira_router_figure_1}), a framework designed to perform zero-shot episodic model routing via natural language task and expert descriptions. Beyond routing, MoIRA is designed for practical specialist serving via adapter-based experts, supporting both disk-based swapping and multi-adapter \cite{predibase2023lorax, sheng2023slora} hot-switching for low-latency deployment. MoIRA sidesteps the scalability constraints of monolithic MoEs by leveraging a pool of pretrained specialists, each finetuned on a focused domain. A lightweight meta-controller dynamically selects the most relevant expert using either embedding-based similarity or prompt-driven reasoning over the textual task specification.

We evaluate MoIRA on two robotic benchmarks: GR1 \cite{nvidia2025gr00tn1openfoundation}, which covers embodiment variation (\texttt{full-body}, \texttt{arms-only}, and \texttt{arms \& waist}), and LIBERO \cite{liu2023libero}, which separates tasks into \texttt{Goal} and \texttt{Spatial} semantic categories. For these experiments, we instantiate MoIRA with the GR00t-N1 and $\pi_0$ VLA backbones, using LoRA adapters \cite{hu2021loralowrankadaptationlarge} to enable efficient specialist training. The routing module is pretrained and frozen, requiring no additional tuning to map tasks to experts.

Our contributions are as follows:
\begin{enumerate}
\item We propose a novel modular routing architecture, MoIRA, that maps tasks to pre-trained experts based on their textual descriptions.
\item We evaluate two routing strategies -- cosine similarity with MiniLM \cite{reimers-2019-sentence-bert} and prompt-based inference with SmolLM2-1.7B \cite{allal2025smollm2smolgoesbig} -- and demonstrate robustness under perturbed inputs.
\item We validate MoIRA on the GR1 and LIBERO benchmarks, showing that it consistently outperforms or achieves parity with generalist models and other MoE approaches on target tasks and previously unseen tasks.
\item We provide an empirical analysis of inference-time expert serving, quantifying the trade-off between VRAM usage and switching latency across (i) fully instantiated adapters, (ii) disk-based swapping, and (iii) multi-LoRA hot-switching for scalable multi-expert deployment.

\end{enumerate}

By decoupling task semantics from execution, MoIRA enables scalable, modular control. It utilizes an architecture-agnostic, external routing mechanism to coordinate a dynamic pool of specialized experts, each implemented as a lightweight LoRA adapter. The zero-shot, language-based routing allows for the independent addition, update, or replacement of experts without costly joint training or retraining the router. It validates a flexible design paradigm for robotic agents that generalize across tasks while benefiting from specialization, providing an alternative to monolithic training pipelines.

\begin{figure}[!h]
  \centering
  \includegraphics[width=\linewidth]{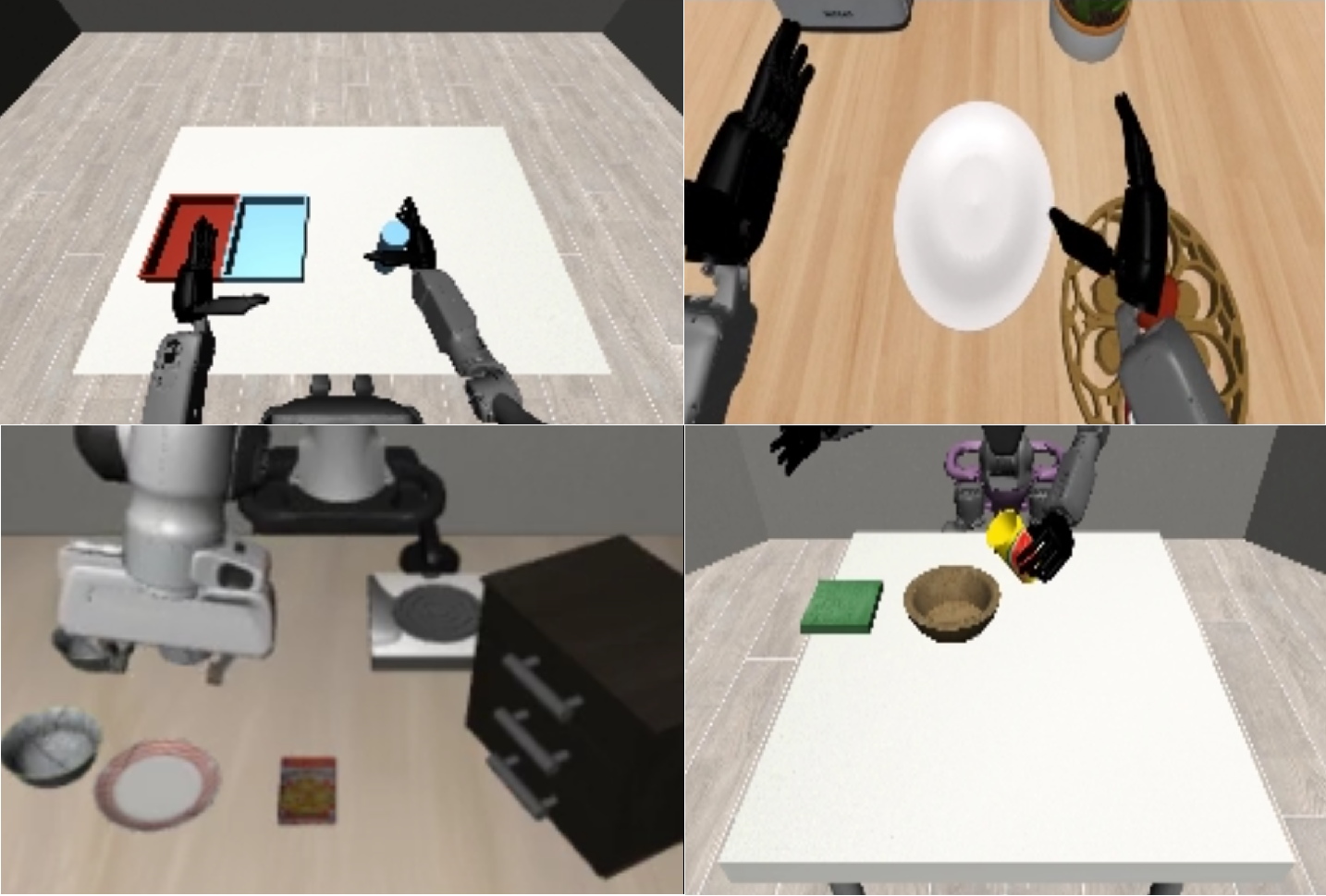}
    \caption{Visualization of policy rollouts. GR1 Humanoid embodiment (\texttt{arms-only}, \texttt{full upper body}) completing pouring and pick-and-place tasks and a Panda Franka arm successfully executing manipulation primitives on the LIBERO \texttt{Spatial} benchmark.}
  \label{fig:rollouts_figure_2}
\end{figure}
\section{Related Work}

\textbf{Policy learning for robotic manipulation.} Classical approaches in robotics rely on RL and IL, both of which have shown strong performance in simulation and real-world environments. RL methods perform well under dense, shaped reward functions, using techniques such as off-policy replay or on-policy actor-critic optimization \cite{sharma2020emergentrealworldroboticskills, chen2021efficientlytrainingonpolicyactorcritic}. However, they often fail in sparse reward regimes due to delayed feedback and exploration bottlenecks. IL methods, particularly behavior cloning \cite{liang2024neverendingbehaviorcloningagentrobotic}, bypass reward engineering by mimicking expert demonstrations, yet suffer from distributional shift and limited generalization when faced with unseen states or perturbations.

\textbf{Foundation models for VLA control.} The emergence of VLMs has shifted the paradigm in robotic perception and planning. Models such as PaLI-Gemma \cite{steiner2024paligemma2familyversatile}, LLaVA \cite{liu2023visualinstructiontuning}, and Qwen-VL \cite{bai2023qwenvlversatilevisionlanguagemodel} demonstrate impressive multimodal grounding capabilities. While not trained for control, these models have inspired a new class of instruction-conditioned control architectures. To bridge the gap between perception and action, recent work extends VLMs into VLA models. These architectures integrate visual-language encoders with control heads, supporting end-to-end policy learning from natural language instructions. Examples include RT-2 \cite{brohan2023rt2visionlanguageactionmodelstransfer} and the RT-X family \cite{embodimentcollaboration2024openxembodimentroboticlearning}. OpenVLA \cite{kim24openvla} and MiniVLA \cite{belkhale2024minivla} focus on smaller, instruction-following backbones, while Physical Intelligence $\pi_0$ \cite{black2024pi0visionlanguageactionflowmodel} and NVIDIA Gr00t-N1 \cite{nvidia2025gr00tn1openfoundation} support scalable deployment through modular control heads. While VLAs are powerful generalists, their monolithic design often results in inefficiencies when scaled to diverse task domains, motivating the development of modular alternatives that preserve generalization while enabling efficient task-level adaptation.

\begin{figure}[!h]
  \centering
  \includegraphics[width=1.0\linewidth]{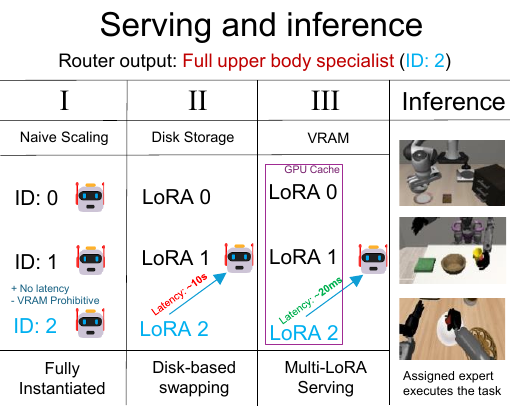}
    \caption{Comparison of expert serving strategies in MoIRA. We illustrate three deployment methods: (I) \textbf{Fully Instantiated} maintains active copies of all experts in memory; while fast, this incurs prohibitive linear VRAM scaling. (II) \textbf{Disk-Based Swapping} reduces memory footprint by loading adapters from storage on-demand, but introduces a high I/O latency bottleneck ($\sim$10s). (III) \textbf{Multi-LoRA Serving} keeps the backbone resident and maintains multiple lightweight adapters concurrently, enabling hot-switching via pointer changes ($\sim$20ms) while avoiding full model replication and large I/O swap costs.}
  \label{fig:moira_inference}
\end{figure}

\textbf{Modularity and Mixture-of-Experts in Robotics.} MoE frameworks provide a natural structure for modularization \cite{6797059, Cai_2025}. However, current robotic MoEs typically rely on \textit{internal} routing mechanisms that are tightly coupled to the model architecture. For instance, MoLe-VLA \cite{zhang2025molevladynamiclayerskippingvision} utilizes a learned Spatial-Temporal Aware Router for dynamic layer skipping, while MoRE \cite{zhao2025moreunlockingscalabilityreinforcement} and GERM \cite{song2024germgeneralistroboticmodel} employ sparse expert activation within shared transformer backbones for locomotion. Unlike these systems, which require joint training and specific internal gating structures, MoIRA treats experts as independent black-box modules, enabling architecture-agnostic coordination.

\textbf{Efficient Inference and Parameter Merging.} Recent work has focused on reducing the computational overhead of MoEs. MoDE \cite{reuss2024efficientdiffusiontransformerpolicies} and Tra-MoE \cite{yang2025tramoelearningtrajectoryprediction} leverage sparsely-gated transformers to optimize multitask policy prediction. Beyond sparse activation, He et al. \cite{he2023merging} demonstrate that merging expert parameters prior to inference can significantly reduce FLOPs and latency without maintaining discrete active experts. While parameter merging effectively lowers computational cost, it assumes that experts share a merge-compatible architecture and does not address the VRAM constraints of hosting diverse, non-mergeable specialist models. MoIRA addresses this distinct hardware constraint via dynamic adapter swapping (Figure \ref{fig:moira_inference}), trading IO latency for the ability to deploy a vast, modular library of LoRA specialists, constrained by disk space rather than VRAM, supporting episodic multi-task deployment on high-capacity hardware. 

Complementary to parameter merging, recent systems work focuses on serving large numbers of LoRA adapters efficiently by keeping a shared backbone resident while multiplexing many low-rank adapters with optimized kernels. Frameworks such as S-LoRA \cite{sheng2023slora} and LoRAX \cite{predibase2023lorax} enable multi-adapter concurrency and millisecond-scale activation, directly addressing the deployment bottleneck in modular expert libraries. MoIRA leverages this multi-LoRA operating point to expand beyond episodic disk swapping while preserving explicit expert boundaries.

\textbf{Semantic Routing and Positioning MoIRA.} Our approach to expert orchestration shares strong methodological parallels with "divide-and-conquer" strategies in NLP, most notably the work of Wang et al. \cite{wang2023divide}. While their framework was developed to decouple data into semantic clusters for zero-shot dialogue state tracking, MoIRA establishes an analogous architecture for the domain of robotic control. Wang et al. apply this separation of concerns to abstract dialogue slots, while MoIRA validates the paradigm's efficacy in handling heterogeneous embodiment and physical task variations. This cross-domain convergence highlights the fundamental robustness of decoupling routing from execution: it ensures that expert policies can be updated, replaced, or added independently without retraining the router, offering a scalable alternative to internal gating that is particularly proven for dynamic real-world environments.
\section{Methods}

MoIRA is a modular meta-controller that selects the best-suited specialist policy based on a task description. It is architecture-agnostic, meaning any expert architecture can be used as a backbone, including but not limited to transformer-based or model-based agents. In our work, we focus on foundation VLAs as expert backbones. The specialists are routed based on their textual description.
Unlike joint or monolithic models, MoIRA functions as an external routing module, enabling flexible expert integration and efficient deployment. 
 
Specialists are routed by MoIRA using embedding-based cosine similarity or prompt-driven language models (Figure \ref{fig:moira_router_figure_3}), with the router requiring no extra training. 

At inference time, MoIRA performs two steps: routing and expert serving (Algorithm~\ref{alg:moira_algo}). Given a task instruction $t$ and expert descriptions $\{m_i\}_{i=1}^K$, it selects the expert index $r$ using either cosine similarity over pretrained text embeddings:
\[
r \leftarrow \arg\max_{i=1 \ldots K} \cos\big(E(t), E(m_i)\big),
\]
or prompt-based reasoning via a frozen LM with Chain-of-Thought \cite{wei2023chainofthoughtpromptingelicitsreasoning} formatting:
\[
r \leftarrow LM_{\text{CoT}} \left([e_1, \ldots, e_n] \parallel t, \{m_i\}\right).
\]
The selected expert $\mathcal{E}_r$ is then served under one of several serving regimes (Figure \ref{fig:moira_inference}) to produce the output trajectory $\tau$ and result $o$.

\begin{algorithm}[!h]
\caption{MoIRA Routing and Serving}
\begin{algorithmic}[1]
\Require Task instruction $t$, Expert pool with meta-descriptions $\{m_i\}_{i=1}^K$
\Ensure Task outcome (trajectory, reward/success signal)

\State \textbf{Routing:}
\If {Embeddings-based routing}
    \State $r \gets \arg\max_{i=1..K} \cos\left(E(t), E(m_i)\right)$ 
\ElsIf {Prompt-driven LM routing}
    \State $r \gets LM_{\text{CoT}}\left([e_1, \ldots, e_n] \, || \, t, \{m_i\}\right)$ 
\EndIf

\State \textbf{Expert serving:}
\If {\textit{All-agents-in-memory}}
    \State Use in-memory model $\mathcal{E}_r$
\Else
    \State Load LoRA adapter weights for $\mathcal{E}_r$
\EndIf
\State Execute task $t$ with expert $\mathcal{E}_r$ to produce trajectory $\tau$ and outcome $o$

\Return $(\tau, o)$
\end{algorithmic}
\label{alg:moira_algo}
\end{algorithm}

The computational complexity of the routing step scales linearly $O(K)$ with the number of experts $K$. Embedding router requires a single forward pass of the encoder $E(t)$ and $K$ dot-product operations, resulting in negligible latency. For the language model router, context length grows linearly with the length of meta-descriptions $\sum_{i=1}^{K} |m_i|$, with an empirical average description length of $\approx 35$ tokens observed in our experiments. While more semantically robust, inference latency increases with the size of the expert pool. 

The primary system bottleneck lies in the expert swap stage. We analyze this constraint through three distinct serving configurations. The simplest, in-memory switching (i), maintains all specialists in VRAM, enabling instantaneous transitions but imposing prohibitive memory requirements that scale linearly with the expert pool. Conversely, disk-based swapping (ii) optimizes for memory by storing inactive experts on disk, though this introduces significant I/O latency during retrieval. To address this trade-off, we consider multi-LoRA serving (iii) by utilizing frameworks such as S-LoRA \cite{sheng2023slora} and LoRAX \cite{predibase2023lorax} as an effective intermediate approach. By leveraging optimized kernels, this method allows for the dynamic swapping of VLA adapters (e.g., 4B parameters) in up to 20ms, effectively balancing low memory footprint with near-real-time responsiveness.

\subsection{Benchmarks and Datasets}

\begin{enumerate}
    \item \textbf{GR1:} We use the GR1 benchmark \cite{nvidia2025gr00tn1openfoundation} containing three embodiments (\texttt{arms-only}, \texttt{arms \& waist}, and \texttt{full upper body}) of GR1 humanoid. Each task contains an annotated task instruction retrieved from \texttt{tasks.jsonl} metadata file.
    \item \textbf{LIBERO:} LIBERO benchmark \cite{liu2023libero} tasks are extracted from \texttt{.bddl} files. We use a modified version of LIBERO splits from \cite{kim24openvla} and select \texttt{Spatial} and \texttt{Goal} task categories of the benchmark as our semantic axis.
\end{enumerate}
Rollout examples from both benchmarks are visualized in Figure \ref{fig:rollouts_figure_2}.

\subsection{Specialist Finetuning}

\begin{figure}[h]
  \centering
  \includegraphics[width=1.0\linewidth]{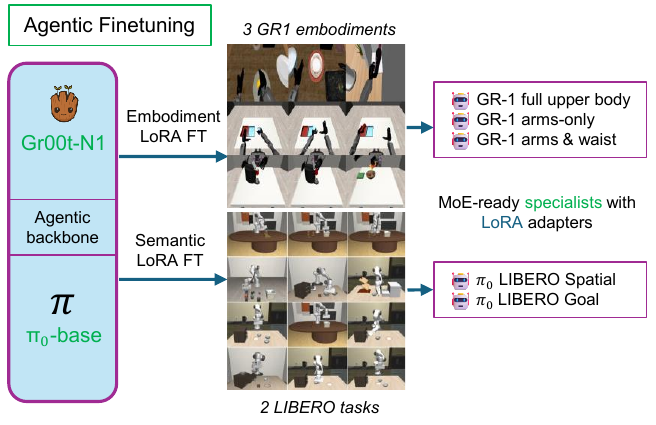}
  \caption{The initial stage of MoIRA's agentic adapter fine-tuning. We finetune VLA backbones (Gr00t-N1, $\pi_0$-base) and derive modular specialists via LoRA: embodiment adapters for GR1 manipulation tasks and semantic adapters for LIBERO \texttt{Goal} and \texttt{Spatial} tasks. Although demonstrated on VLAs, MoIRA’s unified interface can accommodate any agentic backbone, e.g. transformer-based or model-based approaches.}
  \label{fig:method_agentic_finetune_figure_4}
\end{figure}

For GR1, a dedicated GR00t-N1-2B specialist is LoRA fine-tuned for 5K steps for each embodiment. We select three representative tasks to capture variation in embodiment types, task dynamics, and complexity -- \texttt{Pouring}, \texttt{CanSort}, and \texttt{PlacematToPlate}. In addition, we train a jointly-tuned generalist model across all three tasks and reserve \texttt{WineToCabinet} (\texttt{arms \& waist}), \texttt{CuttingboardToTieredBasket} (\texttt{arms \& waist}), and \texttt{Coffee} (\texttt{full upper body}) as previously unseen (held-out) tasks for evaluation.

For LIBERO, we use the $\pi_0$-base-3.3B foundation VLA as the backbone. We create LoRA-adapters for three experts: a spatial task expert, a goal task expert, and a generalist, jointly trained on both tasks -- each model tuned for 30K steps using default hyperparameters. The fine-tuning stage is described in Figure \ref{fig:method_agentic_finetune_figure_4}.

\begin{figure}[!h]
  \centering
  \includegraphics[width=1.0\linewidth]{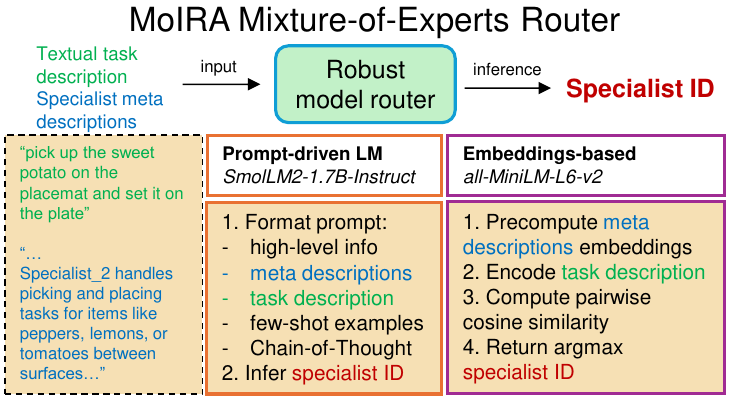}
  \caption{MoIRA MoE routing module. Given a textual task description and specialist meta descriptions, the router assigns a specialist ID via one of two strategies. A prompt-driven language model (SmolLM2) formats all inputs into a single inference prompt with few-shot examples to infer the matching expert (orange). A lightweight embedding-based method (MiniLM) computes cosine similarity between the task and cached meta descriptions to return the closest expert (purple). Both variants support modular, language-based task routing without requiring direct observation input.}
  \label{fig:moira_router_figure_3}
\end{figure}

\subsection{Routing Strategies}

We implement MoIRA task routing using two strategies:

\begin{enumerate}
    \item \textbf{Embedding Similarity:} We embed both task descriptions and expert meta-descriptions using \textit{all-MiniLM-L6-v2} (23M parameters) \cite{reimers-2019-sentence-bert}. The highest cosine similarity among all pairs determines the routed expert ID. This approach treats routing as a geometric retrieval problem, assuming that task instructions and expert capabilities share lexical overlap in a shared latent space. It is computationally negligible but relies on surface-level semantic alignment.
    \item \textbf{Prompt-Driven Routing:} \textit{SmolLM2-1.7B-Instruct} \cite{allal2025smollm2smolgoesbig} receives structured prompts\footnote{SmolLM2 receives a prompt containing meta-descriptions of each expert and a short chain-of-thought instruction set. Example tasks with matched specialists are provided to guide classification. The model is then asked to select the most appropriate expert (e.g., Output: 0, 1, or 2) based on task semantics.}
 containing the task and candidate expert descriptions to select the expert via causal language modeling. Unlike embeddings, this leverages the probabilistic reasoning of a LM to approximate $\texttt{P}(\texttt{Expert}_\texttt{i} | \texttt{Task}, \texttt{Context})$. This allows the router to handle abstract instructions or disparate phrasing where geometric distance may fail, trading inference latency for higher zero-shot routing fidelity.
\end{enumerate}

Each expert is annotated with two distinct types of meta descriptions to support routing:

\begin{enumerate}
\item \textbf{Simple:} short, literal phrases that describe low-level actions or object interactions using task-specific vocabulary (e.g., \texttt{picks and places the black bowl}).
\item \textbf{Abstract:} higher-level, generalized descriptions that emphasize task intent or spatial reasoning without referencing specific objects (e.g., \texttt{transports items between spatial zones}).
\end{enumerate}

This dual-format setup enables compatibility with both routing strategies: the prompt-driven LM benefits from detailed linguistic cues in abstract descriptions, while the embedding-based method relies on surface-level lexical similarity captured in simple ones. It also allows us to evaluate the router’s generalization behavior across levels of semantic abstraction.

\subsection{Classification and Execution}

We assign each task a corresponding ground-truth category: \textbf{embodiment type} (for GR1) or \textbf{semantic type} (for LIBERO). We formulate routing as a multi-class classification problem, where the router predicts the most appropriate specialist given a natural-language task description. Routing accuracy is measured using the macro-averaged F1 score between predicted and ground-truth labels.

At inference time, MoIRA operates in two sequential stages: \textit{routing} and \textit{expert serving}. Given an input task instruction, the router selects an expert index using either embedding-based similarity or prompt-driven language model inference. The selected expert is then served under one of several adapter serving configurations, each exposing a different trade-off between memory footprint and switching overhead.

\begin{itemize}
\item \textbf{Fully instantiated serving.}  
All specialist adapters are kept resident in GPU memory, enabling immediate expert selection. While this configuration minimizes switching overhead, its memory usage scales linearly with the number of specialists, limiting scalability.

\item \textbf{Disk-based swapping.}  
Only a single active adapter is loaded at a time, with inactive specialists stored on CPU or disk and loaded on demand. This approach minimizes GPU memory usage and supports large expert libraries, but incurs significant I/O overhead during expert transitions, making it most suitable for episodic or amortized execution.

\item \textbf{Multi-LoRA serving.}  
An intermediate configuration that maintains multiple lightweight adapters concurrently on a shared backbone, enabling fast expert switching without full model replication. Recent multi-adapter serving frameworks (e.g., S-LoRA \cite{sheng2023slora}, LoRAX \cite{predibase2023lorax}) exemplify this operating point, balancing memory efficiency with low-latency expert activation.
\end{itemize}

These serving strategies allow MoIRA deployments to be adapted to hardware constraints and task dynamics. Disk-based swapping enables scalable, storage-bound expert libraries, while multi-LoRA serving expands the operational envelope toward more reactive, low-latency routing. Crucially, MoIRA preserves explicit expert boundaries and supports heterogeneous, independently trained specialists, in contrast to parameter merging or internally gated MoE architectures. MoIRA serving and inference configurations are summarized in Figure~\ref{fig:moira_inference}.

\subsection{Evaluation Protocol}

For \textbf{GR1}, we report mean squared error (MSE) of the active joints against the expert observations over 50 evaluation trajectories per task. The main joints are \textbf{left hand} and \textbf{right hand} for \texttt{full upper body} and \texttt{arms-only}, and \textbf{right hand} only for \texttt{arms \& waist} embodiment, respectively. 

For \textbf{LIBERO}, we report Success Rate (SR) over 100 rollouts for subtask category (\texttt{Spatial} and \texttt{Goal}). We conduct all our experiments on a single NVIDIA RTX A6000 GPU (48GB VRAM). For all training runs, we keep the default hyperparameters and report evaluation results on 5 seeds.

To ensure robust comparisons, we assess statistical significance using Welch’s t-test for continuous metrics (GR1 MSE) to account for unequal variances between specialists and generalists. For binary success rates (LIBERO), we employ Fisher’s Exact Test. We adopt a standard significance threshold of $p < 0.05$ for all claims.

\section{Results}
For \textbf{GR1}, we analyze specialist finetuning performance, cross-task generalization, instruction robustness, generalization to held-out tasks, and the impact of routing accuracy. 
For \textbf{LIBERO}, we compare specialist vs. generalist performance, contrast MoIRA with prior MoE systems, assess inference-time efficiency, and outline future directions.

\subsection{GR1 Results}

\begin{table}[hbtp]
\centering
\caption{GR1 Specialists vs. Baseline comparison (MSE ± std over 50 Trajectories).}
\begin{tabular}{l|ccc}
\toprule
\textbf{Model} & \textbf{Pouring} & \textbf{CanSort} & \textbf{PlacematToPlate} \\
\midrule
Baseline             & 1.02 ± 0.04   & 0.70 ± 0.10   & 1.67 ± 0.52 \\
Specialist$^*$      & \textbf{0.008 ± 0.004} & \textbf{0.010 ± 0.010} & \textbf{0.310 ± 0.140} \\
\bottomrule
\end{tabular}

\scriptsize {$^*$Each task is evaluated using a target-embodiment specialist (3 total).}
\label{tab:gr1_finetune_table_1}
\end{table}

\subsubsection{Specialist Performance vs. Pretrained Baseline}
We fine-tune each embodiment-specific specialist for 5K steps and compare it with the pretrained GR00t-N1 baseline in Table \ref{tab:gr1_finetune_table_1}. We observe significant reductions in MSE: \texttt{Pouring} (1.02 $\rightarrow$ 0.008, 123× reduction), \texttt{CanSort} (0.70 $\rightarrow$ 0.010, 70× reduction), and \texttt{PlacematToPlate} (1.67 $\rightarrow$ 0.31, 5× reduction). We find the reduction in error to be statistically significant across all tasks ($p < 0.001$ for all 3 tasks), confirming that the reported performance gains are robust to rollout variance and not artifactual.

\begin{table}[!h]
\centering
\caption{Various GR1 cross-task MSE ± std evaluation scenarios using two routing types and simple-abstract meta-descriptions.}
\scriptsize \small
\begin{tabular}{l|ccc}
\toprule
\textbf{Model} & \textbf{Pouring} & \textbf{CanSort} & \textbf{PlacematToPlate} \\
\midrule
Baseline & 1.023 ± 0.036 & 0.699 ± 0.091 & 1.673 ± 0.520 \\
Specialist$^*$     & \textbf{0.008 ± 0.004} & \textbf{0.010 ± 0.010} & \textbf{0.311 ± 0.139} \\
Jointly-tuned & 0.012 ± 0.009 & 0.065 ± 0.010 & 1.187 ± 0.074 \\
\midrule
MiniLM (simple)     & 0.009 ± 0.005 & 0.006 ± 0.005 & 0.336 ± 0.455 \\
MiniLM (abstract)   & 0.008 ± 0.002 & 0.008 ± 0.006 & 0.798 ± 0.614 \\
SmolLM2 (simple)    & 0.008 ± 0.006 & 0.008 ± 0.006 & \textbf{0.241 ± 0.143} \\
SmolLM2 (abstract)  & \textbf{0.009 ± 0.004} & \textbf{0.009 ± 0.007} & 0.287 ± 0.097 \\
\bottomrule
\end{tabular}

\scriptsize{*Each task is evaluated using a target-embodiment specialist (3 total).}
\label{tab:gr1_cross_task_table_2}
\end{table}

\subsubsection{Cross-Task Generalization}
Table \ref{tab:gr1_cross_task_table_2} presents all GR1 task evaluations across specialists and the jointly-tuned generalist. The specialists demonstrated statistically significant reductions in MSE across all evaluated tasks. Specifically, the improvement was significant for \texttt{Pouring} ($p < 0.01$) and highly significant for \texttt{CanSort} and \texttt{PlacematToPlate} ($p < 0.001$). Modular specialists achieve superior precision that is statistically distinguishable from the generalist baseline.


\begin{table}[!h]
\centering
\caption{MoIRA vs.\ Jointly-Tuned Generalist Performance on Held-out GR1 Tasks (MSE $\pm$ std over 10 episodes per task).}
\begin{tabular}{l|ccc}
\toprule
\textbf{Model}                      & \textbf{Wine}         & \textbf{Cuttingboard$^*$}   & \textbf{Coffee}        \\
\midrule
MoIRA      & \textbf{1.55 $\pm$ 0.04} & \textbf{0.41 $\pm$ 0.26} & 0.19 $\pm$ 0.02      \\
Joint Generalist             & 1.87 $\pm$ 0.05     & 1.29 $\pm$ 0.06       & \textbf{0.16 $\pm$ 0.01} \\
\bottomrule
\end{tabular}

\scriptsize{$^*$CuttingboardToTieredBasket.}
\label{tab:moira_vs_generalist_holdout_table_7}
\end{table}

\subsubsection{Robustness to Instruction Perturbations}
To evaluate routing robustness, we semantically perturb\footnote{Perturbed instructions are semantically equivalent rephrasings generated manually for each task, varying lexical choices and syntax (e.g., “\texttt{pick up the pear on the placemat and set it on the plate}” vs. “\texttt{transfer the pear from the mat to the plate}”).}
task instructions by rephrasing them with synonymous verbs and alternate phrasing. For example, \texttt{"take the bell pepper from the placemat and move it to the plate"} may be rewritten as \texttt{"grab the pepper off the placemat and put it onto the plate"}. 

As shown in Table \ref{tab:gr1_perturbed_table_3} and Figure \ref{fig:smollm_mse_figure_6}, SmolLM2 maintains stable performance across tasks regardless of prompt type. In contrast, MiniLM (abstract) shows sharp degradation (e.g. MSE=0.16 on \texttt{CanSort}, Figure \ref{fig:minilm_mse_figure_5} ), indicating higher sensitivity to linguistic variation.

\begin{table}[!h]
\centering
\caption{Router MSE on Original vs. Perturbed Task Descriptions (10 Episodes per Task).}
\begin{tabular}{l|cc|cc|cc}
\toprule
\textbf{Routing} & \multicolumn{2}{c}{\textbf{Pouring}} & \multicolumn{2}{c}{\textbf{CanSort}} & \multicolumn{2}{c}{\textbf{Placemat}} \\
 & Orig. & Pert. & Orig. & Pert. & Orig. & Pert. \\
\midrule
MiniLM simple   & 0.009 & \textbf{0.008} & \textbf{0.006} & 0.009 & \textbf{0.336} & 0.423 \\
MiniLM abstract & 0.008 & 0.008 & \textbf{0.008} & 0.160 & 0.798 & \textbf{0.412} \\
SmolLM2 simple  & 0.008 & \textbf{0.005} & \textbf{0.008} & 0.009 & \textbf{0.241} & 0.293 \\
SmolLM2 abstract & 0.008 & \textbf{0.007} & 0.009 & \textbf{0.008} & \textbf{0.287} & 0.317 \\
\bottomrule
\end{tabular}
\label{tab:gr1_perturbed_table_3}
\end{table}

\subsubsection{Generalization to Held-Out Tasks}
We evaluate MoIRA on three held-out GR1 tasks. MoIRA significantly outperforms a jointly-tuned generalist model on two out of three tasks, achieving lower MSE on \texttt{WineToCabinet} (1.55 against 1.87) and \textsf{CuttingboardToTieredBasket} (0.41 against 1.29), as shown in Table \ref{tab:moira_vs_generalist_holdout_table_7}. On held-out tasks, MoIRA significantly outperformed the generalist on \texttt{WineToCabinet} and \textsf{CuttingboardToTieredBasket} ($p < 0.001$), demonstrating robust zero-shot routing to unseen scenarios. For the \texttt{Coffee} task, the generalist maintained a slight statistical edge ($p < 0.01$), though the absolute difference in MSE was marginal ($0.03$).

\subsubsection{Routing Accuracy and Control Performance}
Accurate routing is essential for effective control. As shown in Table \ref{tab:gr1_heldout_table_4}, MiniLM (simple) achieves perfect routing (F1 = 1.0) with low MSE across tasks. In contrast, SmolLM2 (abstract) yields lower MSE on \textsf{CuttingboardToTieredBasket} (0.735) but suffers from reduced routing reliability (F1 = 0.87). When the routing F1 score is near 1.0, all models yield low MSE. However, control quality deteriorates once routing F1 drops below $\sim$0.9. This trend is most evident in MiniLM (abstract), which frequently misroutes tasks under perturbation. In contrast, SmolLM2, particularly with simple prompts, maintains strong routing fidelity and downstream control performance due to more reliable classification.

\begin{table}[!h]
\centering
\caption{Router F1 Score and Held-out GR1 Tasks Evaluation (MSE $\pm$ std over 10 episodes per task).}
\setlength{\tabcolsep}{4pt}
\begin{tabular}{l|ccc|c}
\toprule
\textbf{Model} & \textbf{WineToCabinet} & \textbf{Cuttingboard$^*$} & \textbf{Coffee} & \textbf{F1 (avg)} \\
\midrule
MiniLM$^a$     & 1.662 ± 0.046 & 0.294 ± 0.209 & 0.196 ± 0.018 & 1.00 \\
MiniLM$^b$   & 1.669 ± 0.047 & 0.399 ± 0.287 & 0.201 ± 0.023 & 1.00 \\
SmolLM2$^a$    & 1.664 ± 0.047 & 0.482 ± 0.413 & 0.197 ± 0.016 & 0.97 \\
SmolLM2$^b$  & 1.669 ± 0.046 & 0.735 ± 0.677 & 0.192 ± 0.019 & 0.87 \\
\bottomrule
\end{tabular}

\scriptsize{$^*$CuttingboardToTieredBasket.\\$^a$ simple meta descriptions; $^b$ abstract meta descriptions}
\label{tab:gr1_heldout_table_4}
\end{table}

\begin{figure}[!h]
  \centering
  \includegraphics[width=1.0\linewidth]{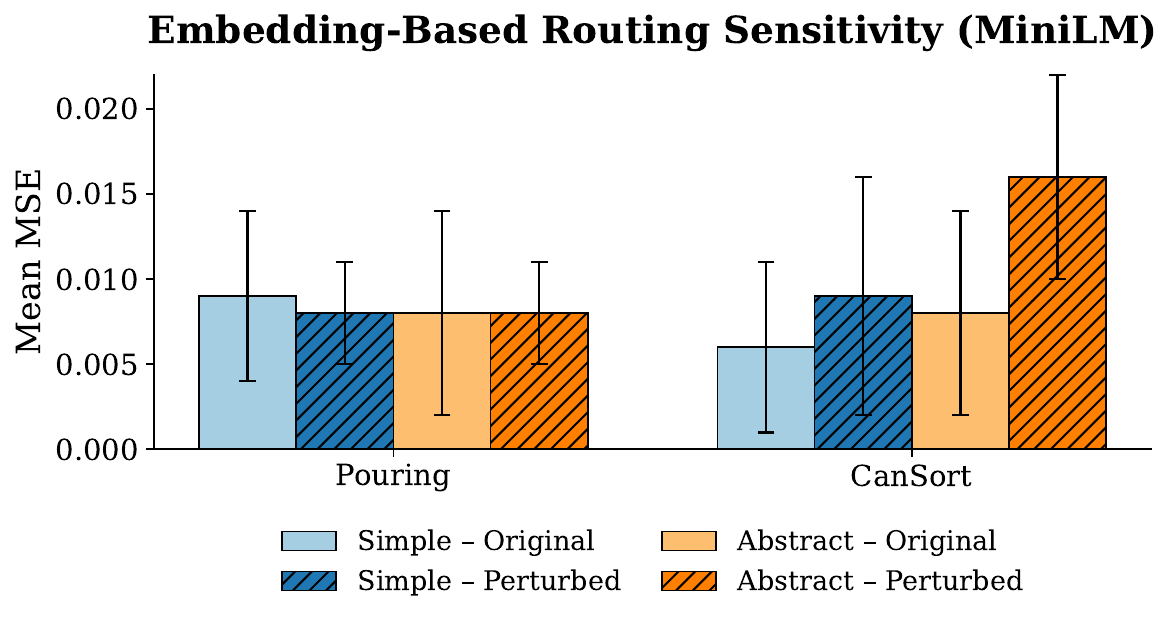}
  \caption{\textbf{MiniLM Routing Sensitivity.} MSE results on GR1 tasks using embedding-based routing. The plot compares performance using simple versus abstract meta-descriptions. While effective on original instructions, the router's performance shows possible degradation on perturbed instructions, particularly when relying on abstract descriptions. This indicates sensitivity to phrasing variations and potential limitations in semantic robustness.}
  \label{fig:minilm_mse_figure_5}
\end{figure}

\begin{figure}[!h]
  \centering
  \includegraphics[width=1.0\linewidth]{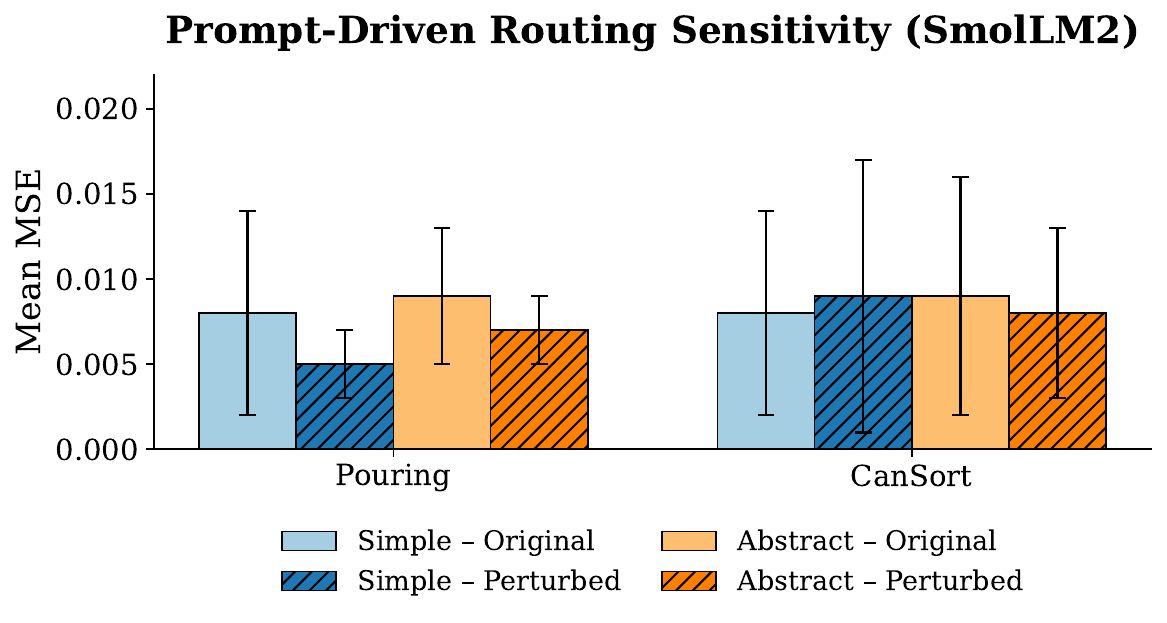}
  \caption{\textbf{SmolLM2 Routing Sensitivity.} MSE results using the prompt-driven language model router. In contrast to the embedding baseline, SmolLM2 maintains low error rates across both original and perturbed instructions for both tasks. This suggests stronger zero-shot reasoning capabilities and stability against linguistic perturbations.}
  \label{fig:smollm_mse_figure_6}
\end{figure}

\subsection{LIBERO Results}

Figure \ref{fig:pi0_generalist_specialist_figure_7} and Table \ref{tab:libero_sr_table_5} report success rates on LIBERO \texttt{Spatial} and \texttt{Goal} subsets. First, we note that the zero-shot pre-trained $\pi_0$ baseline yields 0\% SR, as it lacks the requisite observation normalization statistics to navigate through episodes successfully. To contextualize the performance of MoIRA's specialists, we compare three distinct fine-tuning regimes, trained for 30k steps each:

\begin{enumerate}
    \item \textbf{Global Generalist (Figure \ref{fig:pi0_generalist_specialist_figure_7}, gray bar):} a monolithic baseline fine-tuned on the complete LIBERO dataset (gray bars). This represents the performance upper bound of a strong generalist.
    \item \textbf{Joint LoRA Generalist (Figure \ref{fig:pi0_generalist_specialist_figure_7}, green bar):} a single adapter trained jointly on both \texttt{Spatial} and \texttt{Goal} tasks (green bars), representing a parameter-efficient multi-task baseline.
    \item \textbf{Decoupled Specialists (Figure \ref{fig:pi0_generalist_specialist_figure_7}, blue and yellows bars):} independent LoRA adapters fine-tuned exclusively on their respective subsets (blue and yellow bars), routed via MoIRA.
\end{enumerate}

\begin{table}[!h]
\centering
\caption{LIBERO Specialists and Generalist Success Rates over 100 Rollouts}
\begin{tabular}{l|cc}
\toprule
\textbf{Model} & \textbf{LIBERO Spatial, \%} & \textbf{LIBERO Goal, \%} \\
\midrule
Spatial Specialist & 94 & --- \\
Goal Specialist & --- & \textbf{93} \\
Joint Generalist & \textbf{95} & 90 \\
\bottomrule
\end{tabular}
\label{tab:libero_sr_table_5}
\end{table}

\subsubsection{Specialist vs Generalist Success Rates}
While the jointly-tuned generalist achieves marginally higher raw success rates (95\% vs 94\% \texttt{Spatial}; 90\% vs 93\% \texttt{Goal}), there was no statistically significant difference revealed between the models ($p > 0.05$). Specifically, for \texttt{Spatial} tasks ($p = 1.0$) and \texttt{Goal} tasks ($p \approx 0.61$), the 95\% confidence intervals significantly overlap. These results statistically confirm performance parity -- MoIRA matches the generalist's SR while offering the architectural benefits of modularity and reduced inference memory.

\begin{table}[htpb]
\centering
\caption{Comparison of MoE Flexibility and SR on LIBERO Benchmark. MoIRA outperforms Tra-MoE and achieves parity with MoDE.}
\begin{tabular}{l|ccc}
\toprule
\textbf{Method} & \textbf{Spatial} & \textbf{Goal} & \textbf{Routing} \\
\midrule
Tra-MoE + mask       & 73  & 78  & Internal   \\   
MoDE                 & 90 & 97 & Internal  \\
\textbf{MoIRA (ours)} & \textbf{94} & \textbf{93} & External   \\
\bottomrule
\end{tabular}
\label{tab:libero_general_comp_table_6}
\end{table}

Individual specialists reach 94\% \texttt{Spatial} and 93\% \texttt{Goal} success rate, while the jointly-tuned generalist achieves 95\% and 90\%, respectively. Precise MoE routing is an important aspect of retaining parity with a generalist model.

\subsubsection{Comparison to Prior MoE Systems}
Table \ref{tab:libero_general_comp_table_6} compares MoIRA to Tra-MoE \cite{yang2025tramoelearningtrajectoryprediction} and MoDE \cite{reuss2024efficientdiffusiontransformerpolicies}. MoIRA’s external routing achieves competitive SR while retaining modularity and expert flexibility. In contrast to internal routing (e.g., token-level gating in MoDE), MoIRA’s design enables episodic routing with adaptable backbones.

\begin{figure}[h]
  \centering
  \includegraphics[width=1.0\linewidth]{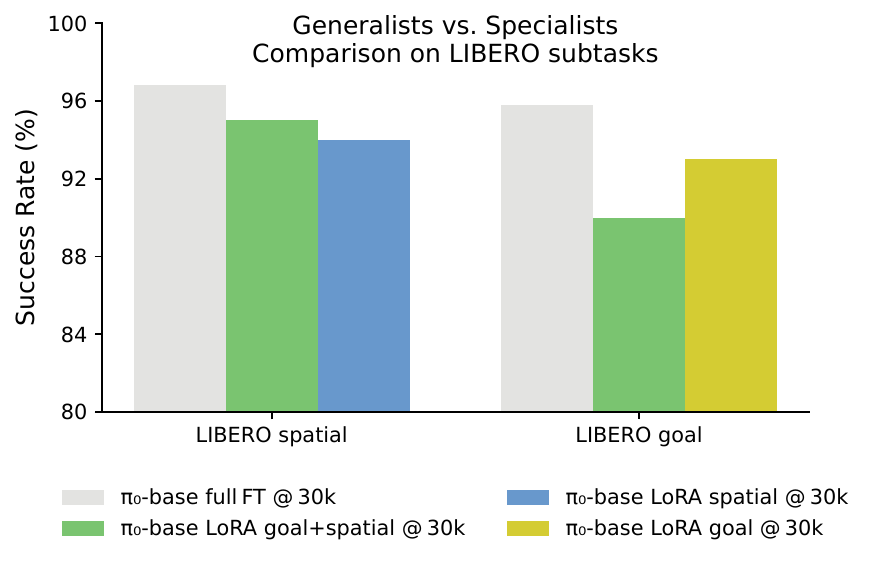}
  \caption{Comparative Success Rates (SR) on LIBERO subtasks. We contrast a \textbf{Global Generalist} (gray; full model fine-tuning on the entire dataset) against a \textbf{Joint LoRA Generalist} (green; single adapter trained on combined Spatial and Goal data) and \textbf{MoIRA-routed Specialists} (blue/yellow; decoupled adapters for each subtask). The modular specialists achieve competitive performance with the monolithic baselines despite being trained on strictly partitioned data.}
  \label{fig:pi0_generalist_specialist_figure_7}
\end{figure}

Reuss et al. \cite{reuss2024efficientdiffusiontransformerpolicies} achieve 90\% SR on LIBERO \texttt{Spatial} and 97\% LIBERO \texttt{Goal} using a sparse expert diffusion transformer. MoIRA scores 93.5\% weighted-average SR across \texttt{Spatial} and \texttt{Goal} subtasks and achieves statistical parity with MoDE (93.5\% vs 93.5\%), matching state-of-the-art diffusion policies using a modular VLA architecture. Unlike MoDE’s internal token-level routing, MoIRA supports episodic routing via fast adapters, making it suitable for deployment-limited settings. 

We also compare our approach with Tra-MoE \cite{yang2025tramoelearningtrajectoryprediction}, which achieves 73\% SR on \texttt{Spatial} and 78\% SR on \texttt{Goal} using a transformer-based policy with internal MoE blocks. MoIRA outperforms Tra-MoE on both suites while offering broader routing flexibility and simpler adapter management.

\subsubsection{Routing Fidelity and System-Level Gains}
Consistent with GR1 findings, routing accuracy in LIBERO strongly correlates with downstream task success. SmolLM2 achieves perfect classification performance (F1=1.0) in a zero-shot manner, enabling specialist experts to reliably outperform the jointly-tuned generalist. This underscores that routing fidelity is the critical point for realizing the full potential of modular expert systems.

\section{Discussion}

This work presents MoIRA, a modular MoE routing framework that dynamically assigns pretrained VLA specialists based on robot embodiment or task semantics. To our knowledge, it is the first framework to demonstrate benchmark-validated dynamic specialization of VLA models along these axes. MoIRA uses LoRA adapters and natural language routing to enable strong end-to-end control without requiring MoE retraining. Our results demonstrate that this modular architecture achieves performance parity with monolithic generalists and other robotic MoE systems while significantly enhancing scalability and deployment efficiency. By decoupling expert specialization from orchestration, MoIRA validates that independent, lightweight adapters offer a flexible and resource-efficient alternative to rigid joint-training pipelines.

\textbf{Sim-to-Real and Sensory Robustness.} 
Current evaluations are performed in simulation. While this allowed us to validate algorithmic stability across highly distinct morphologies (GR1) and semantic domains (LIBERO), real-world deployment introduces actuation noise and sensory drift. The router's reliance on purely textual descriptions currently assumes well-formed instructions and ignores environmental context (e.g., distinguishing between two identical cups). To address this, future work will explore multimodal routing by integrating lightweight vision encoders (e.g., CLIP) or extracting features from the frozen VLA backbone itself. This would allow the router to disambiguate tasks based on visual state without sacrificing the architecture-agnostic nature of the external controller.

\textbf{Latency and Operational Envelope.} 
A key trade-off in our design is the I/O latency introduced by standard dynamic adapter loading. This characteristic typically restricts MoIRA to episodic or batch-processing workflows, such as a service robot completing a 20-minute \textit{"kitchen cleaning"} sequence before switching to \textit{"laundry folding"}, rather than high-frequency, reactive switching. To address this bottleneck, we integrated specialized multi-LoRA serving frameworks, which facilitate concurrent serving of fine-tuned adapters with millisecond-level switching overhead. While this approach effectively masks latency and expands the operational envelope for a subset of compatible backbones, further investigation into universal architecture support and speculative routing strategies remains a priority for fully reactive deployment.

\textbf{Scalability and Automation.} 
Finally, while SmolLM2 demonstrates robust zero-shot routing, the reliance on manually written meta-descriptions limits rapid scaling to thousands of experts. Automating this pipeline is a high-priority research direction. We envision using large language models to auto-generate expert summaries from raw task specifications or demonstration logs, effectively closing the loop to create a fully autonomous, self-organizing mixture of robotic experts.

In summary, MoIRA offers a modular, scalable approach to deploying robotic agents that combine embodiment-specific specialization with semantic task understanding. It demonstrates that pretrained experts can be orchestrated through zero-shot lightweight text-based routing, enabling generalization without relying on joint training or monolithic models.
\endgroup

\bibliographystyle{elsarticle-num-names} 
\bibliography{sample}

\end{document}